\documentclass[pdflatex,sn-nature]{sn-jnl}

\usepackage{graphicx}%
\usepackage{multirow}%
\usepackage{amsmath,amssymb,amsfonts}%
\usepackage{amsthm}%
\usepackage{mathrsfs}%
\usepackage[title]{appendix}%
\usepackage{xcolor}%
\usepackage{textcomp}%
\usepackage{manyfoot}%
\usepackage{booktabs}%
\usepackage{algorithm}%
\usepackage{algorithmicx}%
\usepackage{algpseudocode}%
\usepackage{listings}%
\usepackage{tabularx}
\usepackage{makecell}
\usepackage{array}
\usepackage{threeparttable}
\usepackage{subcaption}
\usepackage{caption}
\usepackage{url}
\usepackage{hyperref}

\theoremstyle{thmstyleone}%
\theoremstyle{thmstyletwo}%
\theoremstyle{thmstylethree}%

\begin{document}

\title[Article Title]{The Gait Signature of Frailty: Transfer Learning based Deep Gait Models for Scalable Frailty Assessment}

\author[1]{\fnm{Laura} \sur{McDaniel}}\email{lmcdan11@jhu.edu}
\equalcont{These authors contributed equally to this work.}

\author[1]{\fnm{Basudha} \sur{Pal}}\email{bpal5@jhu.edu}
\equalcont{These authors contributed equally to this work.}

\author[2]{\fnm{Crystal} \sur{Szczesny}}\email{cszczes1@jh.edu}

\author[1]{\fnm{Yuxiang} \sur{Guo}}\email{yguo87@jhu.edu}

\author[3]{\fnm{Zhaoyang} \sur{Wang}}\email{zwang303@jhu.edu}

\author[4,5]{\fnm{Ryan} \sur{Roemmich}}\email{rroemmi1@jhmi.edu}

\author[2]{\fnm{Peter} \sur{Abadir}}\email{pabadir1@jhmi.edu}

\author*[1,6]{\fnm{Rama} \sur{Chellappa}}\email{rchella4@jhu.edu}

\affil[1]{\orgdiv{Department of Electrical and Computer Engineering},
\orgname{Johns Hopkins University},
\orgaddress{\city{Baltimore}, \state{MD}, \country{USA}}}

\affil[2]{\orgdiv{Division of Geriatrics and Gerontology Medicine},
\orgname{Johns Hopkins University School of Medicine},
\orgaddress{\city{Baltimore}, \state{MD}, \country{USA}}}

\affil[3]{\orgdiv{Department of Computer Science},
\orgname{Johns Hopkins University},
\orgaddress{\city{Baltimore}, \state{MD}, \country{USA}}}

\affil[4]{\orgdiv{Center for Movement Studies},
\orgname{Kennedy Krieger Institute},
\orgaddress{\city{Baltimore}, \state{MD}, \country{USA}}}

\affil[5]{\orgdiv{Department of Physical Medicine and Rehabilitation},
\orgname{Johns Hopkins University School of Medicine},
\orgaddress{\city{Baltimore}, \state{MD}, \country{USA}}}

\affil[6]{\orgdiv{Department of Biomedical Engineering},
\orgname{Johns Hopkins University},
\orgaddress{\city{Baltimore}, \state{MD}, \country{USA}}}

\abstract{Frailty is a key aging condition, yet its assessment remains subjective and difficult to scale. Gait is a sensitive marker of biological aging, but computer vision approaches have been limited by small, imbalanced datasets. We introduce a publicly available silhouette-based frailty gait dataset spanning the full frailty spectrum, including walking-aid users. Using this dataset, we evaluate adaptation of pretrained gait recognition models under limited data. Across architectures, performance depends more on transfer strategy than model complexity: selectively freezing low-level representations while adapting higher-level features yields more stable and generalizable results. Careful handling of class imbalance improves discrimination between clinically adjacent states, and interpretability analyses highlight lower-limb and pelvic regions consistent with frailty biomechanics. These findings support gait-based representation learning as a scalable and interpretable framework for frailty assessment.}


\maketitle

\section{Main}

\begin{figure}[h!]
    \centering
    \includegraphics[width=\linewidth]{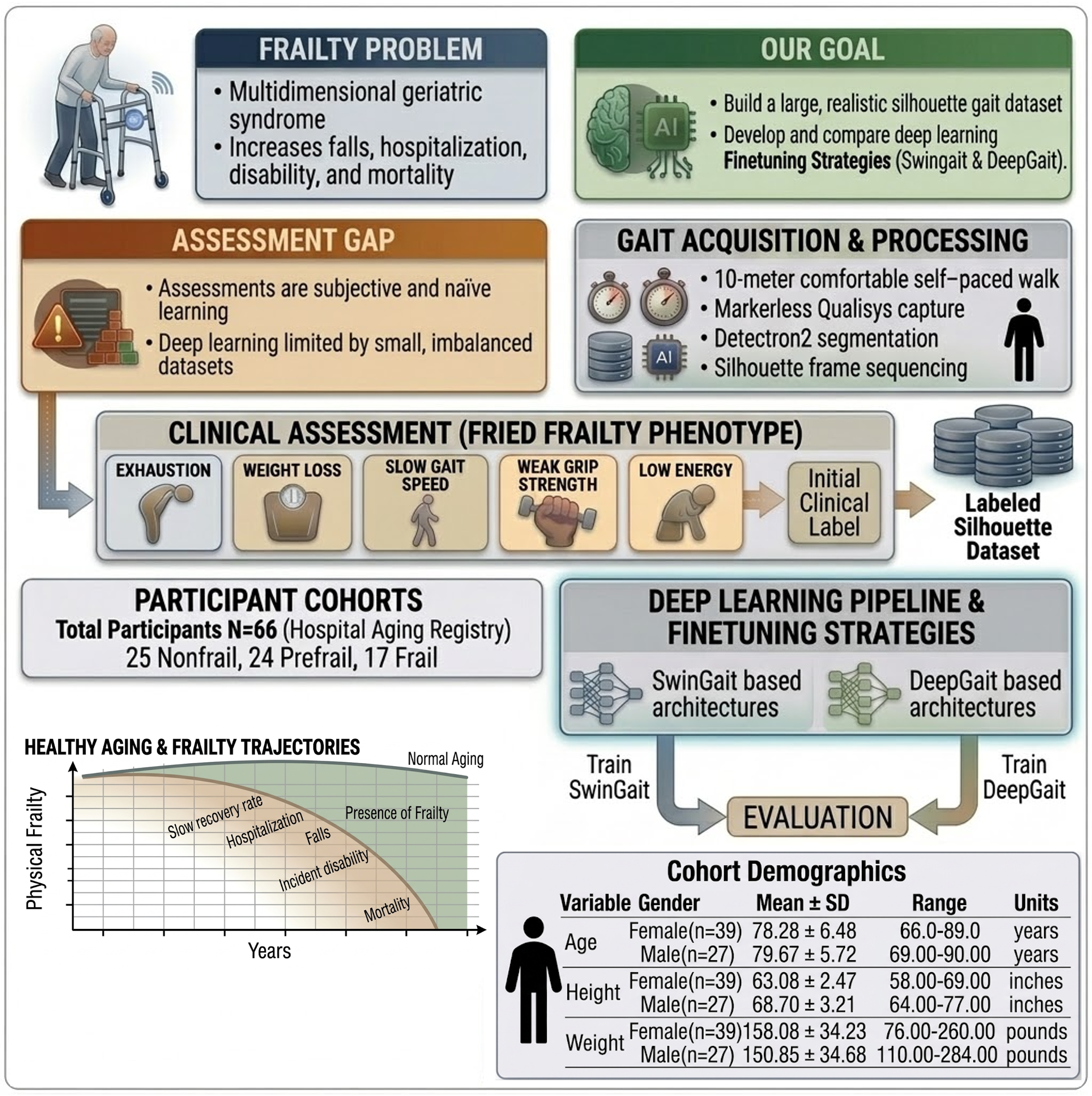}
    \caption{The figure presents the overall framework from clinical motivation to model evaluation. Frailty is introduced as a multidimensional geriatric syndrome motivating objective gait based assessment. Clinical labels are derived using the Fried frailty phenotype defined by five criteria exhaustion, weight loss, slow gait speed, weak grip strength, and low energy, with mobility impairment particularly slow gait speed serving as a core marker of physical frailty. Gait data are collected during a ten meter self paced walk using markerless capture, followed by segmentation and silhouette frame sequencing to construct a labeled silhouette dataset. The conceptual trajectory of age related physical frailty highlights progressive gait deterioration as a primary functional manifestation. The dataset is used to train and compare SwinGait and DeepGait architectures under different finetuning strategies. The bottom panel summarizes demographic characteristics of the study cohort stratified by gender.}
    \label{fig:one}
\end{figure}

Frailty is a multidimensional geriatric syndrome characterized by diminished physiological reserve and increased vulnerability to stressors, leading to heightened risks of falls, hospitalization, disability, institutionalization, cognitive decline, healthcare expenditure, and mortality \cite{Xue2011,Clegg2013,KimRockwood2024,Nari2023,Chi2021}. Frailty is influenced by multiple demographic, biological, and social factors, including age, sex, comorbidity burden, nutritional status, and social isolation \cite{Neri2012,Mello2014,Espinoza2007,Syddall2010}. Although frailty is dynamic and potentially reversible, transitions between robust, prefrail, and frail states vary substantially across individuals and across time \cite{Kojima2019}. These transitions are influenced by underlying health status, comorbidities, and functional decline trajectories \cite{Blaum2005,Chang2012}. With the rapid global expansion of the older population, frailty has emerged as a major public health priority \cite{Cheong2020,Kolle2023,Yi2023}. The worldwide population aged 80 years and above is projected to more than triple between 2015 and 2050, with parallel increases in frailty-related morbidity and health system burden.

Despite its central importance in aging medicine, no universally accepted diagnostic standard exists. Frailty is conceptually distinct from disability and comorbidity, although these conditions frequently overlap in older adults \cite{Fried2004}. Widely used instruments including the Fried Frailty Phenotype \cite{Fried2001}, the Clinical Frailty Scale \cite{Mendiratta2023}, and performance-based measures such as the Timed Up and Go test, differ substantially in thresholds, interpretation, and predictive validity across populations \cite{Jung2020,Park2025,Cesari2014}. Comparative evaluations demonstrate that frailty classifications can diverge depending on the selected tool \cite{Cesari2014}. The Fried phenotype remains the most widely adopted operational definition of physical frailty, comprising unintentional weight loss, weakness, slow walking speed, low physical activity, and exhaustion \cite{Fried2001}. Among these criteria, gait speed consistently emerges as a particularly sensitive marker of multisystem decline (Fig.~\ref{fig:one}).

Gait represents one of the most integrative biomarkers of biological aging. Alterations in gait speed, stride variability, rhythm, and postural control reflect coordinated dysfunction across musculoskeletal, neuromotor, and cardiopulmonary systems and predict downstream disability, dementia, falls, and mortality years before overt clinical events \cite{Studenski2011,Verghese2006,Beauchet2014,Fhon2016}. Conceptually, physical frailty diverges from normal aging along a trajectory marked by progressive mobility deterioration (Fig.~\ref{fig:one}). These observations have motivated growing interest in automated frailty screening.

Recent years have witnessed an expansion of machine learning approaches for frailty detection, prediction, and classification. Systematic reviews indicate that most existing models rely on tabular demographic, clinical, or laboratory variables, often employing conventional supervised classifiers or shallow neural networks \cite{Oliosi2022,Leghissa2023,McDaniel2025}. Community-based studies have applied machine learning to predict frailty status using demographic and explainable clinical features \cite{Isaradech2025} or to forecast longitudinal frailty progression in cohort datasets \cite{Leme2023}. While these approaches demonstrate feasibility, they depend on structured clinical inputs and do not directly model functional manifestations such as gait.

Vision- and sensor-based approaches remain comparatively limited. Early work applied machine vision to classify frailty using controlled gait recordings \cite{Liu2021}, whereas other studies have employed inertial measurement units or upper-extremity motion sensors with deep learning architectures \cite{Amjad2024,Asghari2025}. Although promising, many prior approaches depend on wearable devices or specialized sensing setups and do not systematically evaluate pretrained representation learning architectures. Consequently, their reproducibility across cohorts and their capacity for generalization remain uncertain. The potential of foundation-scale gait models pretrained on large and diverse visual corpora to enable more robust and transferable frailty assessment therefore remains largely unexplored.

The field of computer vision has undergone a fundamental transformation driven by large-scale self-supervised pretraining, including masked image modeling, vision--language contrastive learning \cite{Radford2021}, and transformer-based architectures. Transfer learning has consistently improved performance across medical imaging domains, particularly in small and heterogeneous datasets \cite{Kim2022,Mustafa2021,Salehi2023}. Recent evidence further demonstrates that pretrained visual representations encode rich demographic and anatomical structure \cite{Pal2025SLVH} even without explicit supervision, highlighting both the promise and responsibility of representation learning in clinical aging research.

In gait recognition, state-of-the-art architectures, including GaitSet \cite{Chao2019}, GaitPart \cite{Fan2020}, GaitGL \cite{Lin2021}, Gait3D \cite{Zheng2022}, and transformer-based SwinGait \cite{Fan2023} achieve robust performance under challenging real-world conditions such as viewpoint variation, occlusion, clothing changes, and outdoor environments. Emerging recurrent vision architectures distilled from Vision Transformers further demonstrate that long-range spatiotemporal dependencies can be preserved while substantially improving computational efficiency \cite{Wei2025}. Despite their direct relevance to mobility modeling, these modern representation learning paradigms have been largely unexplored in aging science. A major limiting factor has been the absence of a publicly accessible, clinically grounded silhouette-based frailty gait dataset spanning the full frailty spectrum.

To address this critical gap, we introduce the first publicly available silhouette-based frailty gait dataset collected in a clinically realistic setting. The dataset spans robust, prefrail, and frail states and explicitly includes older adults who use walking aids, enabling modeling of clinically representative mobility patterns. The study cohort reflects realistic age and sex distributions (Fig.~\ref{fig:one}), facilitating translation to real-world aging populations. Using this dataset, we conduct a systematic evaluation of state-of-the-art pretrained gait architectures under diverse transfer learning strategies. We quantify the impact of backbone freezing, class weighting, and lightweight task heads on predictive performance and computational efficiency. We demonstrate that pretrained transformer-based and convolutional gait models substantially outperform conventional task-specific networks in small and imbalanced clinical datasets. Collectively, our findings establish a scalable, non-invasive, and reproducible framework for frailty assessment and create a methodological bridge between gait biometrics and aging science. Overall, our major contributions are:

\begin{itemize}
    \item We release the first publicly accessible, clinically representative silhouette-based frailty gait dataset spanning the full frailty spectrum and including walking aid users.
    \item We systematically evaluate modern pretrained gait architectures under diverse transfer learning strategies, quantifying the influence of backbone freezing and class weighting.
    \item We demonstrate substantial performance gains over conventional convolutional baselines in small and imbalanced clinical datasets.
    \item We establish a direct methodological bridge between biometric gait modeling and clinically meaningful frailty assessment.
\end{itemize}

\section{Results}

We evaluated frailty classification performance using five-fold participant-level cross-validation to ensure robust generalization assessment in this small, clinically realistic, and class-imbalanced cohort. In each fold, approximately 80\% of participants were allocated to the training set and 20\% to a held-out test set, with all gait sequences from a given individual confined to a single partition to prevent identity leakage. Across folds, the training sets comprised 54--55 participants and the test sets 13--14 participants, with stratified allocation to ensure comparable representation of frail, prefrail, and nonfrail categories in each split. This controlled distribution mitigates fold-specific class skew while preserving the ordinal structure of the frailty labels. By evaluating performance across multiple independent participant partitions, cross-validation reduces dependence on any single train--test split and provides a more reliable estimate of out-of-sample generalization, which is particularly important in limited clinical datasets.

\begin{table*}[h]
\centering
\caption{Performance summary under different freezing strategies. Patch Embedding (PE) is treated as a learnable projection layer. Here \checkmark denotes freezing while $\times$ denotes trainable.}
\label{tab:part1_summary}
\resizebox{\textwidth}{!}{
\begin{tabular}{lcccccc}
\toprule
Experiment &
Micro AUC (\%) &
Cohen's Kappa &
CNN &
PE &
Swin S1 &
Swin S2 \\
\midrule
SwinGait M1 & 0.7721 $\pm$ 0.0664 & 0.5923 $\pm$ 0.1128 & $\times$ & $\times$ & $\times$ & $\times$ \\
SwinGait M2 & 0.7790 $\pm$ 0.0910 & 0.6190 $\pm$ 0.1465 & $\checkmark$ & $\times$ & $\times$ & $\times$ \\
SwinGait M3 & 0.7513 $\pm$ 0.0838 & 0.5903 $\pm$ 0.1301 & $\checkmark$ & $\checkmark$ & $\times$ & $\times$ \\
SwinGait M4 & 0.7250 $\pm$ 0.0865 & 0.5103 $\pm$ 0.2126 & $\checkmark$ & $\checkmark$ & $\checkmark$ & $\times$ \\
SwinGait M5 & 0.6316 $\pm$ 0.0713 & 0.2110 $\pm$ 0.1447 & $\checkmark$ & $\checkmark$ & $\checkmark$ & $\checkmark$ \\
\bottomrule
\end{tabular}
}
\end{table*}

\begin{table*}[h]
\centering
\caption{Performance summary for DeepGaitV2 under different backbone freezing strategies. Convolutional layers are indexed from shallow (Layer 0) to deep (Layer 4).}
\label{tab:deepgait_summary}
\resizebox{\textwidth}{!}{
\begin{tabular}{lccc}
\toprule
Experiment & Frozen Layers & Micro AUC (\%) & Cohen's Kappa \\
\midrule
DeepGaitV2 D0 & None            & 0.7348 $\pm$ 0.0548 & 0.5787 $\pm$ 0.0335 \\
DeepGaitV2 D1 & Layer 0         & 0.7788 $\pm$ 0.0312 & 0.5228 $\pm$ 0.0889 \\
DeepGaitV2 D2 & Layers 0--1     & 0.7612 $\pm$ 0.0594 & 0.5657 $\pm$ 0.0764 \\
DeepGaitV2 D3 & Layers 0--2     & 0.7407 $\pm$ 0.0306 & 0.4947 $\pm$ 0.1241 \\
DeepGaitV2 D4 & Layers 0--3     & 0.7458 $\pm$ 0.0623 & 0.5308 $\pm$ 0.0584 \\
DeepGaitV2 D5 & Layers 0--4     & 0.6961 $\pm$ 0.0567 & 0.4412 $\pm$ 0.0811 \\
\bottomrule
\end{tabular}
}
\end{table*}

\subsection{Effect of backbone freezing strategies}

We investigated the impact of backbone adaptation strategies on frailty classification performance using both a hybrid convolutional--transformer architecture (SwinGait) and a convolutional baseline (DeepGaitV2). Performance was evaluated using five-fold cross-validation and quantified using Micro-AUC and linearly weighted Cohen's Kappa (Tables~\ref{tab:part1_summary} and~\ref{tab:deepgait_summary}). These analyses were designed to determine how varying degrees of representation adaptation influence robustness and ordinal agreement in a small, clinically realistic cohort.

For SwinGait, five configurations (M1--M5) corresponding to progressively more restrictive freezing strategies were evaluated. These configurations varied in whether the convolutional backbone, patch embedding (PE), and Swin Transformer stages (S1, S2) were frozen or fine-tuned. Selective freezing yielded superior performance compared to both full fine-tuning and extensive freezing. Freezing only the convolutional backbone while allowing the patch embedding and transformer stages to adapt (M2) achieved the strongest overall performance, with a Micro-AUC of $0.7790 \pm 0.0910$ and a Cohen's Kappa of $0.6190 \pm 0.1465$, indicating substantial agreement between predicted and true frailty stages and representing the highest ordinal agreement across configurations. Full fine-tuning (M1) produced comparable but slightly lower performance (Micro-AUC $0.7721 \pm 0.0664$; Kappa $0.5923 \pm 0.1128$, corresponding to moderate agreement), indicating that unrestricted adaptation does not confer clear benefit in limited clinical data settings. Freezing both the convolutional backbone and patch embedding (M3) reduced performance (Micro-AUC $0.7513 \pm 0.0838$; Kappa $0.5903 \pm 0.1301$, also reflecting moderate agreement), while further freezing of Swin stage S1 (M4) resulted in a more pronounced decline (Micro-AUC $0.7250 \pm 0.0865$; Kappa $0.5103 \pm 0.2126$, indicating reduced but still moderate agreement). Fully freezing all backbone components except the classification head (M5) led to substantial performance degradation (Micro-AUC $0.6316 \pm 0.0713$; Kappa $0.2110 \pm 0.1447$, corresponding to only fair agreement), demonstrating that static pretrained representations alone are insufficient to capture cohort-specific frailty patterns. Across SwinGait configurations, Micro-AUC values ranged from $0.6316$ to $0.7790$, while Kappa ranged from fair to substantial agreement, underscoring the importance of controlled backbone adaptation for achieving clinically meaningful ordinal reliability.

These trends suggest that the convolutional stem of SwinGait captures low-level silhouette contours and motion primitives that transfer effectively across tasks and should be preserved, whereas the hierarchical transformer stages encode higher-level relational and spatiotemporal dependencies that require adaptation to represent frailty-specific gait alterations. Freezing only the stem therefore preserves stable low-level filters while permitting task-specific refinement of global motion structure, which may explain the superior ordinal agreement observed in M2.

To assess freezing effects within a purely convolutional architecture, we conducted a structured ablation using DeepGaitV2. Convolutional layers were indexed from shallow (Layer 0) to deep (Layer 4), and progressively deeper layers were frozen while the remaining layers were fine-tuned. Selective freezing of early layers yielded the strongest discriminative performance. Freezing only the first convolutional layer (D1) achieved the highest Micro-AUC ($0.7788 \pm 0.0312$), while freezing the first two layers (D2) resulted in slightly lower Micro-AUC ($0.7612 \pm 0.0594$) but improved ordinal agreement, with a Kappa of $0.5657 \pm 0.0764$, indicating moderate agreement. In contrast, progressively freezing deeper portions of the backbone (D3--D5) led to consistent performance degradation. Micro-AUC declined from $0.7407 \pm 0.0306$ (D3) to $0.6961 \pm 0.0567$ (D5), accompanied by a corresponding reduction in Kappa from $0.4947 \pm 0.1241$ to $0.4412 \pm 0.0811$. Fully freezing all convolutional layers (D5) produced the weakest performance overall. Notably, full fine-tuning (D0) did not provide the best results (Micro-AUC $0.7348 \pm 0.0548$; Kappa $0.5787 \pm 0.0335$), suggesting that partial constraint of early feature adaptation may improve generalization in small cohorts.

For DeepGaitV2, freezing only the shallowest convolutional layer (D1) likely preserves generic edge and motion detectors learned during large-scale pretraining while introducing a mild regularization effect that mitigates overfitting in this small cohort. In contrast, full fine-tuning may permit excessive adaptation of early filters, reducing generalization, whereas extensive freezing constrains higher-level representation learning.

Across both architectures, intermediate freezing strategies consistently outperformed both full fine-tuning and extensive freezing. Clinically, these findings indicate that reliable frailty classification from gait silhouettes requires preservation of foundational movement representations while permitting targeted adaptation to population-specific characteristics.

\begin{table*}[h!]
\centering
\caption{Transfer-learning comparison between SwinGait and DeepGaitV2 under selected best freezing strategies with and without inverse square-root class weighting. For SwinGait, the ``Frozen CNN'' configuration corresponds to M2 (CNN frozen, patch embedding and both Swin stages trainable). For DeepGaitV2, the partially frozen configuration corresponds to freezing convolutional Layer 0.}
\label{tab:part4_fair_comparison}
\resizebox{\textwidth}{!}{
\begin{tabular}{llp{4.2cm}cccc}
\toprule
Backbone & Configuration & Frozen Components & Class Weights & Micro AUC & Cohen's Kappa \\
\midrule
SwinGait & Frozen CNN (M2)    & CNN frozen; PE, S1, S2 trainable   & No  & 0.7790 $\pm$ 0.0910 & 0.6190 $\pm$ 0.1465 \\
SwinGait & Frozen CNN (M2)    & CNN frozen; PE, S1, S2 trainable   & Yes & 0.7683 $\pm$ 0.0858 & 0.5658 $\pm$ 0.0824 \\
SwinGait & Unfrozen CNN       & All backbone components trainable  & No  & 0.7721 $\pm$ 0.0664 & 0.5923 $\pm$ 0.1128 \\
SwinGait & Unfrozen CNN       & All backbone components trainable  & Yes & 0.7631 $\pm$ 0.0685 & 0.6091 $\pm$ 0.0981 \\
\midrule
DeepGaitV2 & Partially Frozen (D1) & Layer 0 frozen & No  & 0.7788 $\pm$ 0.0312 & 0.5228 $\pm$ 0.0889 \\
DeepGaitV2 & Partially Frozen (D1) & Layer 0 frozen & Yes & 0.7852 $\pm$ 0.0558 & 0.5248 $\pm$ 0.0765 \\
DeepGaitV2 & Unfrozen              & None           & No  & 0.7348 $\pm$ 0.0548 & 0.5787 $\pm$ 0.0335 \\
DeepGaitV2 & Unfrozen              & None           & Yes & 0.7510 $\pm$ 0.0415 & 0.5155 $\pm$ 0.0967 \\
\bottomrule
\end{tabular}
}
\end{table*}

\subsection{Influence of class weighting under optimized training settings}

To evaluate the impact of class weighting under clinically optimized backbone configurations, we compared models trained with and without inverse-square class weighting across both SwinGait and DeepGaitV2 (Table~\ref{tab:part4_fair_comparison}). In all experiments reported in Table~\ref{tab:part4_fair_comparison}, models were trained using the joint cross-entropy and triplet loss formulation adopted from the original OpenGait framework, thereby isolating the effect of class weighting while holding the loss design constant.

For SwinGait with frozen CNN components (M2), the unweighted model achieved a Micro-AUC of $0.7790 \pm 0.0910$ and a Kappa of $0.6190 \pm 0.1465$. Introducing inverse-square weighting resulted in slightly lower discrimination (Micro-AUC $0.7683 \pm 0.0858$) and reduced ordinal agreement (Kappa $0.5658 \pm 0.0824$). A similar pattern was observed in the fully trainable SwinGait configuration, where weighting marginally decreased Micro-AUC while producing comparable Kappa values. These findings indicate that SwinGait performance is largely stable to moderate imbalance handling but does not derive systematic benefit from inverse-square weighting under participant-level stratification.

For DeepGaitV2, the matched partially frozen configuration corresponded to freezing Layer~0 (D1). In this setting, the unweighted model achieved a Micro-AUC of $0.7788 \pm 0.0312$ and a Kappa of $0.5228 \pm 0.0889$. Applying inverse-square weighting modestly increased Micro-AUC to $0.7852 \pm 0.0558$ while producing only minimal change in Kappa ($0.5248 \pm 0.0765$). In the fully trainable DeepGaitV2 configuration, weighting increased Micro-AUC from $0.7348 \pm 0.0548$ to $0.7510 \pm 0.0415$ but was associated with a reduction in Kappa. Thus, while DeepGaitV2 exhibited small discrimination gains under weighting, ordinal agreement did not consistently improve.

Overall, inverse-square class weighting did not produce consistent improvements across architectures or freezing strategies. Several factors may explain this stability. First, participant-level stratified cross-validation ensured comparable class proportions across folds, reducing variance attributable to fold-specific imbalance. Second, class imbalance in this cohort was moderate rather than extreme, limiting the benefit of aggressive reweighting. Third, the joint triplet plus cross-entropy objective already encourages structured embedding separation across frailty stages, partially mitigating frequency-driven bias in representation learning.

Notably, under the empirically best-performing freezing configuration for each architecture, SwinGait and DeepGaitV2 achieved comparable discriminative performance, with Micro-AUC values in the 0.77--0.79 range. However, SwinGait consistently demonstrated stronger ordinal agreement, with Kappa values exceeding 0.60 in the best configuration, compared to approximately 0.52 for DeepGaitV2. From a translational perspective, these results indicate that architectural sophistication alone does not guarantee superior performance in small clinical cohorts, and that careful backbone adaptation may exert greater influence than imbalance correction or attention-based design.

\subsection{Stage-specific classification performance across frailty categories}

To further evaluate clinical relevance, we examined classification performance for each frailty stage. Across all architectures and freezing strategies, a consistent pattern emerged: the Frail class was identified most reliably, followed by the Nonfrail class, while the Prefrail class was comparatively more difficult to distinguish. This behavior reflects the clinical continuum of frailty. Individuals classified as frail often demonstrate pronounced gait abnormalities, including reduced stride length, slower walking speed, postural instability, and in some cases use of assistive devices. Nonfrail individuals, by contrast, display preserved mobility and stable gait cycles. Prefrail individuals represent a transitional stage characterized by subtle reductions in movement quality and heterogeneous functional decline.

Importantly, most misclassifications occurred between adjacent frailty stages rather than between extremes. In practical terms, the models tended to confuse prefrail individuals with either nonfrail or frail participants, but rarely misidentified clearly frail individuals as fully nonfrail. This preserves the ordinal progression of frailty severity and is reflected in the relatively strong Cohen's Kappa values observed across configurations. From a clinical perspective, such errors are less consequential than confusion between frail and nonfrail categories, which remained well separated.

Stage-wise trends were consistent across backbone freezing strategies. In SwinGait, the best-performing configuration maintained reliable identification of frail individuals while providing reasonable discrimination of nonfrail participants, whereas prefrail classification remained comparatively challenging. More restrictive freezing reduced the model's ability to capture subtle intermediate-stage gait changes. DeepGaitV2 demonstrated similar behavior, with partially frozen configurations achieving strong detection of frailty but reduced differentiation of prefrailty across strategies.

\subsection{Clinical interpretation}

Across all experiments, the most consistent determinant of performance was the transfer-learning strategy rather than architectural complexity or imbalance correction. Both architectures achieved Micro-AUC values in the 0.77--0.79 range under optimal configurations, indicating good overall discrimination of frailty status from gait silhouettes. SwinGait demonstrated stronger ordinal agreement, suggesting improved preservation of graded frailty severity.

From a clinical perspective, reliable identification of frail individuals is particularly important, as this group is at highest risk for adverse outcomes including falls, hospitalization, and functional decline. The lower accuracy for prefrailty reflects biological heterogeneity and aligns with the conceptualization of prefrailty as a transitional stage. Importantly, most errors occurred between neighboring stages, meaning that individuals were typically misclassified by one level of severity rather than across extremes. This behavior preserves clinically meaningful ordering and supports potential use in screening and monitoring scenarios.

These findings support the feasibility of silhouette-based gait analysis for automated frailty screening. Controlled backbone adaptation improves generalization, transformer-based architectures enhance ordinal staging, and class weighting provides limited additional benefit.

\section{Discussion}

In this study, we demonstrate that clinically meaningful frailty classification can be achieved from silhouette-based gait sequences using transfer learning, even within a small and heterogeneous cohort. Frailty is strongly associated with falls, hospitalization, disability, and mortality, yet routine screening remains underutilized due to time, staffing, and infrastructure constraints. By leveraging participant-level five-fold cross-validation, we provide a conservative and robust estimate of generalization performance, reducing reliance on any single train--test partition.

A central finding of this study is that controlled backbone adaptation is essential for stable frailty classification. Across both SwinGait and DeepGaitV2, selectively freezing low-level feature extractors while permitting higher-level adaptation consistently yielded superior performance. This pattern suggests that foundational silhouette and motion primitives learned during large-scale pretraining transfer effectively across tasks, whereas deeper layers require refinement to capture subtle gait alterations associated with frailty, such as reduced stride length, altered cadence, and impaired lower-limb coordination. Fully freezing pretrained backbones degraded performance, indicating that generic biometric gait embeddings alone are insufficient to capture clinically relevant frailty signatures.

Importantly, the consistency of trends across both transformer-based and convolutional architectures suggests that the observed principles are architecture-agnostic. Although SwinGait offers increased modeling flexibility through attention mechanisms, DeepGaitV2 achieved comparable discrimination when paired with an appropriate transfer-learning strategy. These findings indicate that careful representation adaptation may be more influential than architectural sophistication in small-sample clinical settings, highlighting the importance of model-centric optimization when data are limited. This perspective complements emerging data-centric approaches that emphasize dataset quality, reliability, and bias mitigation, which have been shown to substantially improve performance in heterogeneous clinical datasets even without major architectural changes \cite{pal2025grasp}. Together, these observations suggest that future frailty modeling efforts may benefit from jointly exploring model-centric representation refinement and data-centric curation strategies. More broadly, they demonstrate that gait recognition architectures originally developed for biometric person identification can be repurposed to capture clinically meaningful aging phenotypes.

\begin{figure}[h]
    \centering
    \includegraphics[width=\linewidth]{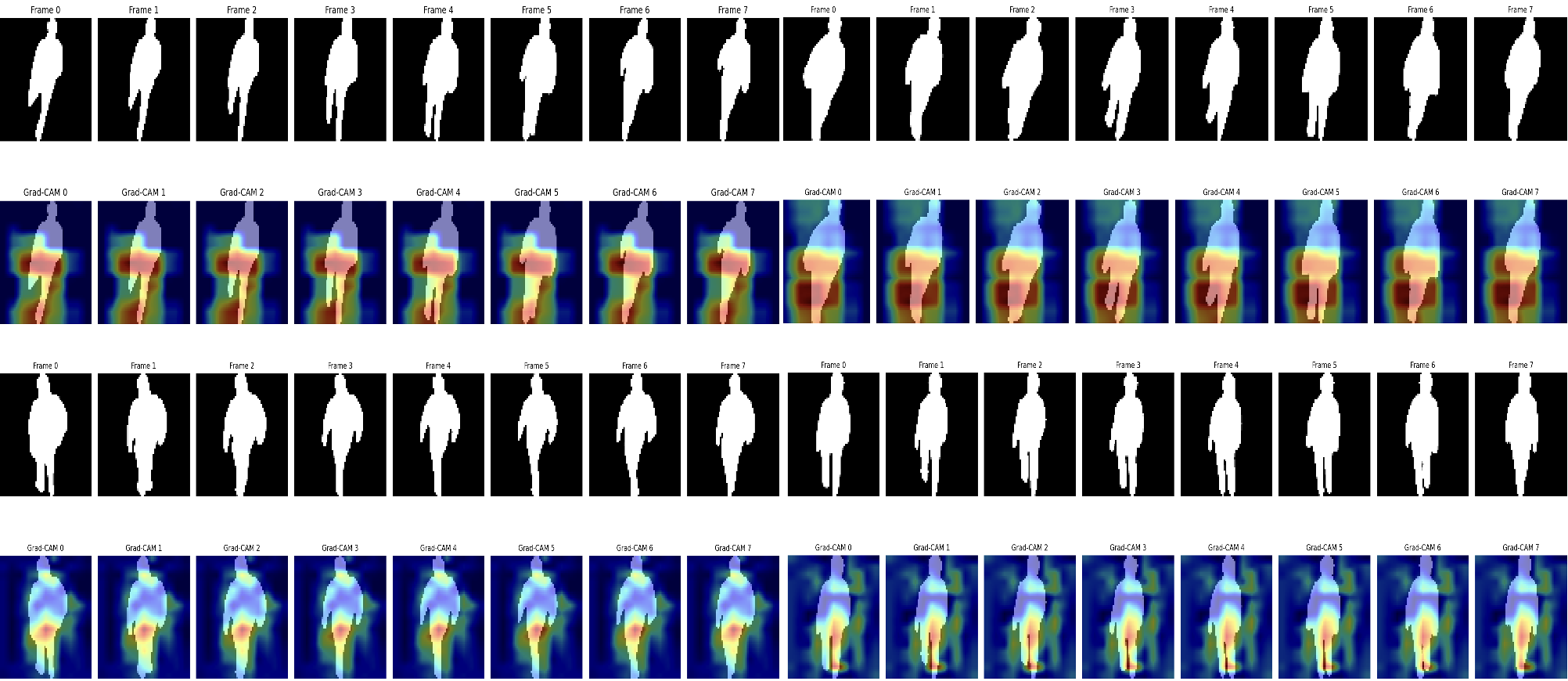}
    \caption{Exemplar of gradient activation maps for best performing models. \textbf{Rows 1 and 2}: SwinGait M2 and its corresponding GradCAM. \textbf{Rows 3 and 4}: DeepGait D1 and its corresponding GradCAM.}
    \label{fig:opt_gradcam}
\end{figure}

From a translational perspective, the use of cross-validation reframes model evaluation around reproducibility and stability rather than isolated peak metrics. In frailty screening contexts where decisions may inform referral, fall-risk mitigation, or rehabilitation planning, consistent ordinal agreement is more valuable than sporadic high accuracy. The interpretability analysis further strengthens clinical plausibility. Grad-CAM visualizations \cite{selvaraju} as seen in Fig.~\ref{fig:opt_gradcam} consistently highlighted the lower extremities across temporal frames, aligning with established geriatric understanding that frailty manifests through diminished lower-limb strength, reduced propulsion, and altered gait dynamics.

Beyond high-resource tertiary centers, silhouette-based gait assessment may hold particular promise in low- and middle-income countries (LMICs) and community-based care settings. Prior work on artificial intelligence implementation in LMIC healthcare systems has largely focused on imaging modalities such as echocardiography, CT, and MRI, demonstrating substantial potential while highlighting infrastructure and scalability challenges \cite{Marey2025LMIC,Mollura2020}. In contrast, automated screening tools for geriatric syndromes such as frailty remain underexplored despite the growing burden of aging populations. Silhouette-based gait analysis offers a privacy-preserving, low-cost alternative that requires only standard RGB video acquisition and modest computational resources.

Several limitations warrant consideration. First, the cohort size remains modest and derived from a single center, limiting immediate generalizability. Although cross-validation mitigates internal partition bias, external validation across diverse populations, walking environments, and recording conditions is essential. Second, while silhouette representations enhance privacy and align with pretrained gait backbones, they omit fine-grained kinematic information that may further refine frailty assessment. Third, the present study focuses on cross-sectional staging rather than longitudinal modeling of frailty progression. Future work will extend this framework to larger, multi-center cohorts and assess external validation across varied demographic and geographic settings. Integration of multimodal data streams, including wearable sensors or clinical covariates, may enhance predictive sensitivity.

In summary, this work establishes a clinically grounded framework for adapting pretrained gait recognition models to frailty classification under limited data conditions. By emphasizing reproducible evaluation, controlled backbone adaptation, and interpretable decision analysis, our findings support the development of scalable, privacy-preserving gait-based tools for frailty screening and global clinical translation.

\section{Methods}

\begin{figure}[h]
    \centering
    \includegraphics[width=\linewidth]{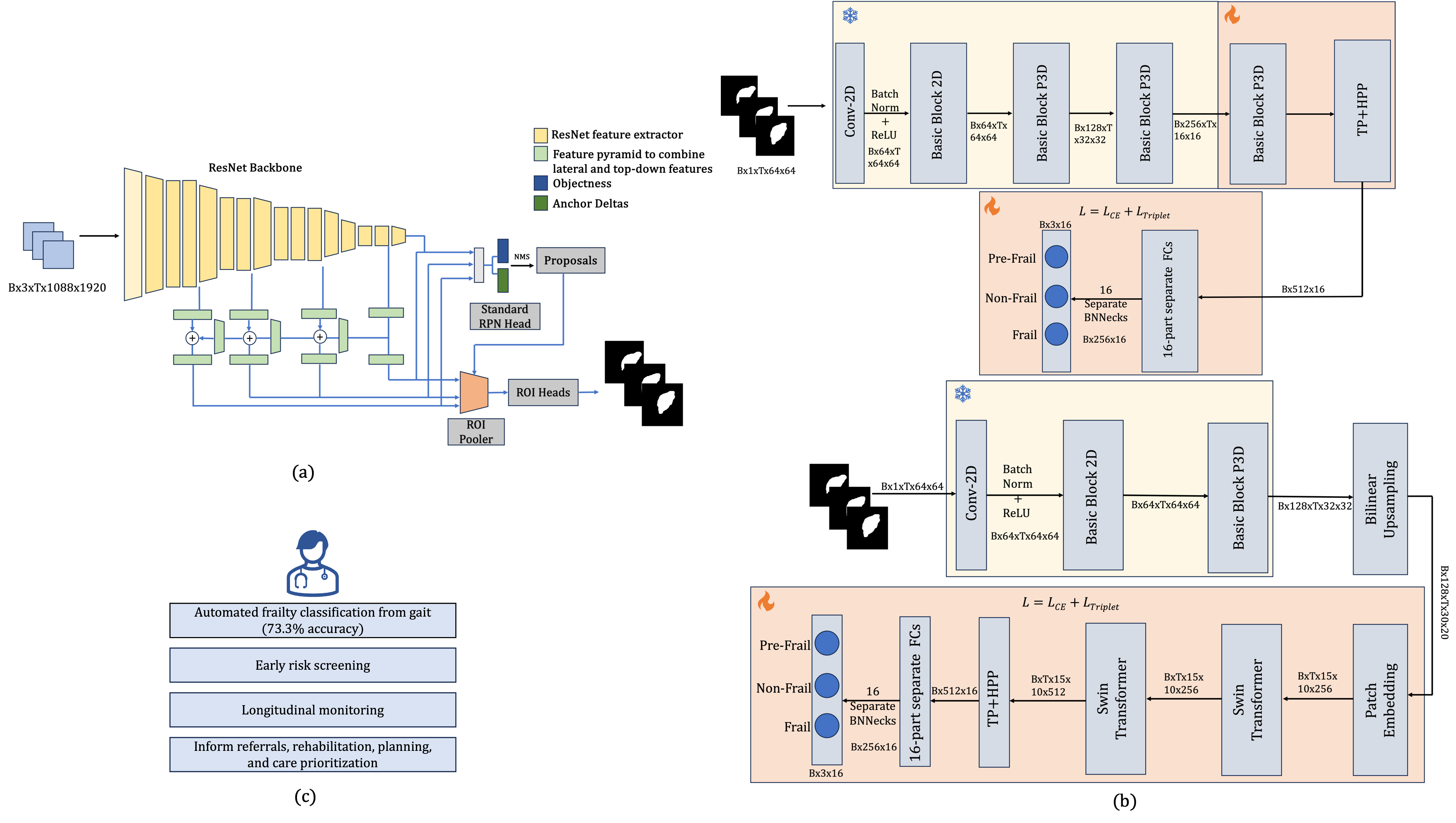}
    \caption{Overview of the proposed frailty assessment framework.
(a) In-clinic RGB videos are converted to privacy-preserving silhouette sequences using Detectron2.
(b) Two transfer-learning backbones, DeepGaitV2 and SwinGait, extract gait representations for frailty prediction.
In DeepGaitV2, the Basic Block 2D consists of two spatial convolution layers with batch normalization and ReLU activation, followed by an identity shortcut,
while the Basic Block P3D replaces full 3D convolution with a pseudo-3D design composed of spatial and temporal convolutions, optional downsampling,
synchronized batch normalization, and a 3D residual shortcut.
(c) Clinician-facing interpretation of the predicted frailty status and potential downstream applications, including screening, longitudinal monitoring, and care planning.}
    \label{fig:teasermain}
\end{figure}

\subsection{Study cohort and data preprocessing}

The study cohort comprised 68 participants recruited from a single academic medical center, representing a spectrum of frailty severity. All participants with complete and usable gait recordings were included in the final analysis. Frailty status was assessed prior to data collection using the Fried Frailty Phenotype, a validated clinical instrument based on five criteria reflecting physiological reserve and vulnerability. Participants with a score of 0 were classified as non-frail, those with scores of 1 or 2 as prefrail, and those with scores of 3 to 5 as frail. The final cohort consisted of 25 non-frail, 24 prefrail, and 17 frail participants. Demographic characteristics of the study cohort, stratified by gender, are summarized in Fig.~\ref{fig:one}.

All participants completed a standardized walking task at their self-selected pace and were permitted to use customary assistive walking devices to ensure safety and ecological validity. Eligibility criteria required the ability to ambulate continuously for at least five minutes and either no neurological diagnosis or a diagnosis of Parkinson disease, stroke, or cerebellar ataxia. Exclusion criteria included congestive heart failure, peripheral artery disease with claudication, active cancer, pulmonary or renal failure, unstable angina, uncontrolled hypertension, dementia, severe aphasia, asthma interfering with oxygen consumption testing, and orthopedic or pain conditions that could limit safe participation.

Gait data were captured using a multi-camera markerless motion capture system comprising eight Arqus cameras and eight Miqus cameras mounted around the perimeter of the walking area. To ensure participant privacy and enable deidentification, raw RGB frames were not used directly for model training. Instead, silhouette representations were extracted from video frames using a pretrained instance segmentation model implemented in Detectron2 \cite{Wu2019Detectron2}. This representation removes identifiable facial and appearance information while preserving body shape and spatiotemporal motion cues relevant to gait analysis. To select an optimal camera view for silhouette extraction, videos from a representative participant were processed across all camera viewpoints. Detection rates ranged from 52\% to 100\%, with silhouette counts varying from 1,105 to 3,762 frames per recording. The camera achieving the highest detection rate and silhouette yield was selected for all subsequent analyses. Silhouette masks were temporally aligned with the original video frames and resized to a fixed spatial resolution. Frames with segmentation failure or severe occlusion were excluded.

The dataset was partitioned at the participant level to prevent subject overlap across splits. Model evaluation was performed using five-fold participant-level cross-validation with approximately 80\% of participants used for training and 20\% held out for testing in each fold. Due to the cohort size, the exact number of participants varied slightly across folds, with training sets containing 54--55 participants and corresponding test sets containing 13--14 participants. Class distributions were preserved across folds, with each training split including approximately 19--20 nonfrail, 20 prefrail, and 15--16 frail participants, and each test split including approximately 4---5 nonfrail, 5 prefrail, and 3--4 frail participants. For training, 60 frames were sampled per participant, while 80 frames per participant were used for testing within each fold.

In addition to privacy considerations, silhouette representations were selected to ensure architectural compatibility. Both backbones used in this study were pretrained for biometric gait recognition, which conventionally operates on silhouette-based inputs. Adopting silhouettes therefore ensured alignment between the pretraining domain and the downstream frailty assessment task, facilitating stable transfer learning under limited data conditions.

\subsection{Overall model framework}

All models followed a common framework consisting of a pretrained gait recognition backbone and a task-specific frailty classification head. Given a sequence of silhouette frames, the backbone learned spatiotemporal gait representations that were aggregated into a fixed-length feature vector and passed to a lightweight classifier trained to predict frailty status. Two pretrained backbones were evaluated: SwinGait, a hybrid convolutional transformer architecture, and DeepGaitV2, a deep convolutional architecture. Both backbones have been validated on large-scale gait recognition benchmarks and were selected for their ability to model subtle motion patterns under unconstrained walking conditions. Figure \ref{fig:teasermain} outlines the overall workflow setup.

\subsection{SwinGait backbone}

SwinGait is a hybrid convolutional transformer architecture designed to efficiently model spatiotemporal dependencies in silhouette-based gait data. A convolutional stem first extracts local spatial features and performs patch embedding, reducing the computational burden associated with applying self-attention to sparse silhouette inputs. Features are then processed through hierarchical Swin Transformer stages composed of shifted window attention blocks. Self-attention is restricted to local windows, while periodic window shifting enables information exchange across windows, allowing the model to capture both local and long-range spatial dependencies.

Spatial resolution is progressively reduced through patch merging operations, while channel dimensionality increases to enhance representational capacity. Temporal dynamics are captured through aggregation across frame-level representations, enabling the model to encode gait characteristics such as cadence, stride regularity, and inter-limb coordination.

\subsection{DeepGaitV2 backbone}

DeepGaitV2 is a deep convolutional architecture designed to capture spatiotemporal gait patterns through hierarchical residual learning. The backbone consists of an initial convolutional stem followed by four residual stages, with channel dimensionality increasing progressively from 64 to 512. Residual blocks may incorporate two-dimensional, three-dimensional, or pseudo-3D convolutions, enabling flexible modeling of spatial and temporal structure.

Spatial downsampling is performed using stride-based convolutions in intermediate stages, while deeper layers capture higher-level motion patterns. Partial freezing of early convolutional layers was explored to retain low-level spatiotemporal filters learned during gait pretraining while allowing deeper layers to adapt to frailty-specific motion cues.

\subsection{Backbone adaptation strategies}

To study how transfer-learning design choices influence frailty classification, we systematically varied backbone adaptation strategies while keeping the overall training pipeline fixed. Across all experiments, the frailty classification head was identical and trained from scratch, while the degree of fine-tuning applied to the pretrained backbone was controlled through selective freezing of backbone components.

\subsubsection{SwinGait backbone freezing strategies}

For SwinGait, we evaluated five backbone adaptation strategies (M1--M5) that progressively increase the degree of freezing across the convolutional stem, patch embedding layer, and Swin Transformer stages. The patch embedding (PE) layer was treated as a learnable projection and could be either frozen or fine-tuned depending on the configuration. A summary of these strategies is provided in Table~\ref{tab:part1_summary}. The configurations are defined as follows:

\begin{itemize}
    \item \textbf{M1 (Fully Fine-Tuned Backbone).} All backbone components, including the convolutional stem, patch embedding layer, and both Swin Transformer stages, were fine-tuned jointly with the classification head. This represents the least restrictive setting.
    \item \textbf{M2 (Frozen CNN Stem Only).} The convolutional stem was frozen, while the patch embedding layer and both Swin Transformer stages were fine-tuned. This setup preserves low-level spatial feature extraction while allowing higher-level representations to adapt to frailty-related gait patterns.
    \item \textbf{M3 (Frozen CNN + Patch Embedding).} The convolutional stem and patch embedding layer were frozen, while both Swin Transformer stages were fine-tuned.
    \item \textbf{M4 (Frozen CNN + PE + Swin Stage 1).} The convolutional stem, patch embedding layer, and Swin Transformer Stage~1 were frozen, while Swin Stage~2 was fine-tuned.
    \item \textbf{M5 (Fully Frozen Backbone).} All backbone components were frozen. Only the classification head was trained. This configuration evaluates direct transferability of pretrained representations without backbone adaptation.
\end{itemize}

\subsubsection{DeepGaitV2 backbone freezing strategies}

For DeepGaitV2, backbone adaptation was explored by freezing progressively deeper convolutional stages while fine-tuning the remaining layers. Convolutional layers were indexed from shallow (Layer~0) to deep (Layer~4), following the original DeepGaitV2 architecture. The evaluated configurations (D0--D5), summarized in Table~\ref{tab:deepgait_summary}, are defined as follows:

\begin{itemize}
    \item \textbf{D0 (Fully Fine-Tuned).} No convolutional layers were frozen; the entire backbone and classification head were fine-tuned end-to-end.
    \item \textbf{D1 (Freeze Layer 0).} The shallowest convolutional layer (Layer~0) was frozen, while Layers~1--4 were fine-tuned.
    \item \textbf{D2 (Freeze Layers 0--1).} The first two convolutional layers were frozen.
    \item \textbf{D3 (Freeze Layers 0--2).} Early and mid-level convolutional layers were frozen.
    \item \textbf{D4 (Freeze Layers 0--3).} All but the deepest convolutional layer were frozen.
    \item \textbf{D5 (Fully Frozen Backbone).} All convolutional layers (Layers~0--4) were frozen, and only the classification head was trained.
\end{itemize}

\subsection{Frailty classification head}

For both backbones, output feature maps were aggregated using temporal pooling across frames followed by horizontal pooling across spatial partitions to preserve region-specific motion cues. The resulting fixed-length representation was passed to a frailty classification head consisting of fully connected layers with batch-normalization bottlenecks and a final three-class output corresponding to non-frail, prefrail, and frail categories.

\subsection{Optimization and training details}

All models were trained using the AdamW optimizer with a learning rate of $1 \times 10^{-4}$ and a weight decay of $1 \times 10^{-3}$. Training was performed with a batch size of 4 and input sequences consisting of 60 frames sampled with a frame skip of 3. Model evaluation was conducted using five-fold participant-level cross-validation. For each fold, models were trained independently and evaluated every 500 training iterations on the fold-specific held-out test set, for a total of 10{,}000 training iterations per fold. Training employed a combination of cross-entropy loss and triplet loss. For class-weighted configurations, inverse square-root class weights were applied to the cross-entropy loss to address class imbalance. Data augmentation was applied uniformly across all configurations and included random affine transformations, horizontal flipping, perspective distortion, silhouette cutting, random rotation, and dropout.

\subsection{Evaluation metrics}

Model performance was evaluated using the area under the receiver operating characteristic curve (AUROC) and linearly weighted Cohen's Kappa. AUROC quantifies the ability of the model to distinguish between frailty categories across all possible classification thresholds. For multiclass evaluation, a micro-averaged formulation was used to aggregate performance across classes. Cohen's Kappa measures agreement between predicted and true labels while accounting for agreement expected by chance. In this study, we report linearly weighted Cohen's Kappa to account for the ordered structure of frailty categories (nonfrail, prefrail, frail). Linear weighting assigns progressively larger penalties to more severe misclassifications, ensuring that the metric reflects clinically meaningful ordinal disagreement.

Performance was computed independently within each fold of the five-fold participant-level cross-validation framework. Final model performance is reported as the mean and standard deviation across folds. No model selection was performed based on test-set performance within folds; instead, cross-validation served to quantify expected out-of-sample behavior under fixed training configurations.

\section*{Data Availability}

The de-identified silhouette dataset along with the metadata information will be publicly available and can be downloaded from this link: \url{https://drive.google.com/drive/folders/1V1GM4XeteDnSa1MSmj7o45ZvU_9CjQnJ?usp=sharing}

\section*{Code Availability}

The code is available at: \url{https://github.com/lauramcdaniel006/CF_OpenGait}
\section*{Author Contributions}
L.M. and B.P. contributed equally to this work. L.M. contributed to conceptualization, clinical study design, data curation, investigation, software implementation, formal analysis and writing of the original draft. B.P. led conceptualization of the computational framework, methodology development, software implementation, formal analysis, validation, visualization, interpretation of clinical relevance and writing of the original draft. C.S. contributed to data curation, clinical investigation, and validation of frailty assessments. Y.G. assisted with software development and methodological support. Z.W. contributed to proofreading and technical review. R.R. provided expertise in gait biomechanics, contributing to conceptualization and clinical interpretation of movement-related findings along with funding acquisition. P.A. contributed to conceptualization, clinical supervision, resources, and interpretation of aging-related outcomes. R.C. contributed to conceptualization, supervision, and overall guidance of methodology and manuscript preparation. All authors reviewed and approved the final manuscript.

\section*{Funding}

R.R. received funding from the Claude D. Pepper Older Americans Independence Center for supporting this research (National Institute on Aging grant P30-AG021334). 

\section*{Ethics Declarations}

\textbf{Ethics approval and consent to participate:}
This study was approved by the Johns Hopkins Institutional Review Board 
(approval number: 00255175). All participants provided written informed 
consent prior to participation.

\section*{Competing Interest}
The authors declare no competing interests.
\nocite{*}
\bibliography{references}

@article{Xue2011,
  title={The frailty syndrome: definition and natural history},
  author={Xue, Qian-Li},
  journal={Clinics in geriatric medicine},
  volume={27},
  number={1},
  pages={1},
  year={2011}
}

@article{Clegg2013,
  title={Frailty in elderly people},
  author={Clegg, Andrew and Young, John and Iliffe, Steve and Rikkert, Marcel Olde and Rockwood, Kenneth},
  journal={The lancet},
  volume={381},
  number={9868},
  pages={752--762},
  year={2013},
  publisher={Elsevier}
}

@article{KimRockwood2024,
  title={Frailty in older adults},
  author={Kim, Dae Hyun and Rockwood, Kenneth},
  journal={New England Journal of Medicine},
  volume={391},
  number={6},
  pages={538--548},
  year={2024},
  publisher={Mass Medical Soc}
}

@article{Nari2023,
  title={Impact of frailty on mortality and healthcare costs and utilization among older adults in South Korea},
  author={Nari, Fatima and Park, Eun-Cheol and Nam, Chung-Mo and Jang, Sung-In},
  journal={Scientific reports},
  volume={13},
  number={1},
  pages={21203},
  year={2023},
  publisher={Nature Publishing Group UK London}
}

@article{Chi2021,
  title={Impacts of frailty on health care costs among community-dwelling older adults: a meta-analysis of cohort studies},
  author={Chi, Junting and Chen, Fei and Zhang, Jing and Niu, Xiaodan and Tao, Hongxia and Ruan, Haihui and Wang, Yanhong and Hu, Junping},
  journal={Archives of gerontology and geriatrics},
  volume={94},
  pages={104344},
  year={2021},
  publisher={Elsevier}
}

@article{Neri2012,
  title={Relationships between gender, age, family conditions, physical and mental health, and social isolation of elderly caregivers},
  author={Neri, Anita Liberalesso and Yassuda, M{\^o}nica Sanches and Fortes-Burgos, Andr{\'e}a Cristina Garofe and Mantovani, Efig{\^e}nia Passarelli and Arbex, Fl{\'a}via Silva and de Souza Torres, Stella Vidal and Perracini, M{\^o}nica Rodrigues and Guariento, Maria Elena},
  journal={International Psychogeriatrics},
  volume={24},
  number={3},
  pages={472--483},
  year={2012},
  publisher={Cambridge University Press}
}

@article{Mello2014,
  title={Health-related and socio-demographic factors associated with frailty in the elderly: a systematic literature review},
  author={Mello, Amanda de Carvalho and Engstrom, Elyne Montenegro and Alves, Luciana Correia},
  journal={Cadernos de saude publica},
  volume={30},
  pages={1143--1168},
  year={2014},
  publisher={SciELO Public Health}
}

@article{Fried2004,
  title={Untangling the concepts of disability, frailty, and comorbidity: implications for improved targeting and care},
  author={Fried, Linda P and Ferrucci, Luigi and Darer, Jonathan and Williamson, Jeff D and Anderson, Gerard},
  journal={The journals of Gerontology Series A: Biological sciences and Medical sciences},
  volume={59},
  number={3},
  pages={M255--M263},
  year={2004},
  publisher={Oxford University Press}
}

@article{Espinoza2007,
  title={Risk factors for frailty in the older adult},
  author={Espinoza, Sara E and Fried, Linda P},
  journal={Clinical Geriatrics},
  volume={15},
  number={6},
  pages={37},
  year={2007},
  publisher={MULTIMEDIA HEALTHCARE FREEDOM LLC}
}

@article{Syddall2010,
  title={Prevalence and correlates of frailty among community-dwelling older men and women: findings from the Hertfordshire Cohort Study},
  author={Syddall, Holly and Roberts, Helen C and Evandrou, Maria and Cooper, Cyrus and Bergman, Howard and Sayer, Avan Aihie},
  journal={Age and ageing},
  volume={39},
  number={2},
  pages={197--203},
  year={2010},
  publisher={Oxford University Press}
}

@article{Blaum2005,
  title={The association between obesity and the frailty syndrome in older women: the Women's Health and Aging Studies},
  author={Blaum, Caroline S and Xue, Qian Li and Michelon, Elisabete and Semba, Richard D and Fried, Linda P},
  journal={Journal of the American Geriatrics Society},
  volume={53},
  number={6},
  pages={927--934},
  year={2005},
  publisher={Wiley Online Library}
}

@article{Chang2012,
  title={Frailty and its impact on health-related quality of life: a cross-sectional study on elder community-dwelling preventive health service users},
  author={Chang, Yaw-Wen and Chen, Wei-Liang and Lin, Fu-Gong and Fang, Wen-Hui and Yen, Ming-Yung and Hsieh, Chia-Chuan and Kao, Tung-Wei},
  journal={PloS one},
  volume={7},
  number={5},
  pages={e38079},
  year={2012},
  publisher={Public Library of Science San Francisco, USA}
}

@article{Kojima2019,
  title={Transitions between frailty states among community-dwelling older people: a systematic review and meta-analysis},
  author={Kojima, Gotaro and Taniguchi, Yu and Iliffe, Steve and Jivraj, Stephen and Walters, Kate},
  journal={Ageing research reviews},
  volume={50},
  pages={81--88},
  year={2019},
  publisher={Elsevier}
}

@article{Cheong2020,
  title={Risk factors of progression to frailty: findings from the Singapore longitudinal ageing study},
  author={Cheong, CY and Nyunt, MSZ and Gao, Q and Gwee, X and Choo, RWM and Yap, KB and Wee, Shiou Liang and Ng, Tze-Pin},
  journal={The Journal of nutrition, health and aging},
  volume={24},
  number={1},
  pages={98--106},
  year={2020},
  publisher={Elsevier}
}

@article{Kolle2023,
  title={Reversing frailty in older adults: a scoping review},
  author={Kolle, Aur{\'e}lie Tonjock and Lewis, Krystina B and Lalonde, Michelle and Backman, Chantal},
  journal={BMC geriatrics},
  volume={23},
  number={1},
  pages={751},
  year={2023},
  publisher={Springer}
}

@article{Yi2023,
  title={Healthy aging, early screening, and interventions for frailty in the elderly},
  author={Deng, Yi and Zhang, Keming and Zhu, Jiali and Hu, Xiaofeng and Liao, Rui},
  journal={Bioscience trends},
  volume={17},
  number={4},
  pages={252--261},
  year={2023},
  publisher={International Research and Cooperation Association for Bio \& Socio-Sciences~…}
}

@article{Fried2001,
  title={Frailty in older adults: evidence for a phenotype},
  author={Fried, Linda P and Tangen, Catherine M and Walston, Jeremy and Newman, Anne B and Hirsch, Calvin and Gottdiener, John and Seeman, Teresa and Tracy, Russell and Kop, Willem J and Burke, Gregory and others},
  journal={The journals of gerontology series a: biological sciences and medical sciences},
  volume={56},
  number={3},
  pages={M146--M157},
  year={2001},
  publisher={Oxford University Press}
}

@article{Jung2020,
  title={Screening value of timed up and go test for frailty and low physical performance in Korean older population: the Korean Frailty and Aging Cohort Study (KFACS)},
  author={Jung, Hee-Won and Kim, Sunyoung and Jang, Il-Young and Shin, Dong Wook and Lee, Ji Eun and Won, Chang Won},
  journal={Annals of geriatric medicine and research},
  volume={24},
  number={4},
  pages={259},
  year={2020}
}

@article{Park2025,
  title={Comparative Analysis of Three Frailty Assessment Tools: A Cross-Sectional Study},
  author={Park, DaSol},
  journal={INQUIRY: The Journal of Health Care Organization, Provision, and Financing},
  volume={62},
  pages={00469580251363877},
  year={2025},
  publisher={SAGE Publications Sage CA: Los Angeles, CA}
}

@article{Cesari2014,
  title={Frailty: an emerging public health priority},
  author={Cesari, Matteo and Prince, Martin and Thiyagarajan, Jotheeswaran Amuthavalli and De Carvalho, Islene Araujo and Bernabei, Roberto and Chan, Piu and Gutierrez-Robledo, Luis Miguel and Michel, Jean-Pierre and Morley, John E and Ong, Paul and others},
  journal={Journal of the American Medical Directors Association},
  volume={17},
  number={3},
  pages={188--192},
  year={2016},
  publisher={Elsevier}
}

@article{Studenski2011,
  title={Gait speed and survival in older adults},
  author={Studenski, Stephanie and Perera, Subashan and Patel, Kushang and Rosano, Caterina and Faulkner, Kimberly and Inzitari, Marco and Brach, Jennifer and Chandler, Julie and Cawthon, Peggy and Connor, Elizabeth Barrett and others},
  journal={Jama},
  volume={305},
  number={1},
  pages={50--58},
  year={2011},
  publisher={American Medical Association}
}

@article{Verghese2006,
  title={Quantitative gait dysfunction and risk of cognitive decline and dementia},
  author={Verghese, Joe and Wang, Cuiling and Lipton, Richard B and Holtzer, Roee and Xue, Xiaonan},
  journal={Journal of Neurology, Neurosurgery \& Psychiatry},
  volume={78},
  number={9},
  pages={929--935},
  year={2007},
  publisher={BMJ Publishing Group Ltd}
}

@article{Beauchet2014,
  title={Gait analysis in demented subjects: Interests and perspectives},
  author={Beauchet, Olivier and Allali, Gilles and Berrut, Gilles and Hommet, Caroline and Dubost, V{\'e}ronique and Assal, Fr{\'e}d{\'e}ric},
  journal={Neuropsychiatric disease and treatment},
  volume={4},
  number={1},
  pages={155--160},
  year={2008},
  publisher={Taylor \& Francis}
}

@article{Fhon2016,
  title={Fall and its association with the frailty syndrome in the elderly: systematic review with meta-analysis},
  author={Fhon, Jack Roberto Silva and Rodrigues, Rosalina Aparecida Partezani and Neira, Wilmer Fuentes and Huayta, Violeta Magdalena Rojas and Robazzi, Maria Lucia do Carmo Cruz},
  journal={Revista da Escola de Enfermagem da USP},
  volume={50},
  number={06},
  pages={01005--01013},
  year={2016},
  publisher={SciELO Brasil}
}

@article{Oliosi2022,
  title={Machine learning approaches for the frailty screening: a narrative review},
  author={Oliosi, Eduarda and Guede-Fern{\'a}ndez, Federico and Londral, Ana},
  journal={International Journal of Environmental Research and Public Health},
  volume={19},
  number={14},
  pages={8825},
  year={2022},
  publisher={MDPI}
}

@article{Leghissa2023,
  title={Machine learning approaches for frailty detection, prediction and classification in elderly people: A systematic review},
  author={Leghissa, Matteo and Carrera, {\'A}lvaro and Iglesias, Carlos A},
  journal={International journal of medical informatics},
  volume={178},
  pages={105172},
  year={2023},
  publisher={Elsevier}
}

@article{Isaradech2025,
  title={Machine learning models for frailty classification of older adults in Northern Thailand: Model development and validation study},
  author={Isaradech, Natthanaphop and Sirikul, Wachiranun and Buawangpong, Nida and Siviroj, Penprapa and Kitro, Amornphat},
  journal={JMIR aging},
  volume={8},
  pages={e62942},
  year={2025},
  publisher={JMIR Publications Toronto, Canada}
}

@article{Leme2023,
  title={Machine learning models to predict future frailty in community-dwelling middle-aged and older adults: the ELSA cohort study},
  author={Leme, Daniel Eduardo da Cunha and De Oliveira, Cesar},
  journal={The Journals of Gerontology: Series A},
  volume={78},
  number={11},
  pages={2176--2184},
  year={2023},
  publisher={Oxford University Press US}
}

@article{Liu2021,
  title={Application of machine vision in classifying gait frailty among older adults},
  author={Liu, Yixin and He, Xiaohai and Wang, Renjie and Teng, Qizhi and Hu, Rui and Qing, Linbo and Wang, Zhengyong and He, Xuan and Yin, Biao and Mou, Yi and others},
  journal={Frontiers in Aging Neuroscience},
  volume={13},
  pages={757823},
  year={2021},
  publisher={Frontiers Media SA}
}

@article{Amjad2024,
  title={Deep Learning for Frailty Classification Using IMU Sensor Data: Insights From FRAILPOL Database},
  author={Amjad, Arslan and Szcz{\k{e}}sna, Agnieszka and B{\l}aszczyszyn, Monika and Sacha, Jerzy and Sacha, Magdalena and Feusette, Piotr and Wola{\'n}ski, Wojciech and Konieczny, Mariusz and Borysiuk, Zbigniew and Khan, Basheir},
  journal={IEEE Sensors Journal},
  volume={25},
  number={2},
  pages={3974--3981},
  year={2024},
  publisher={IEEE}
}

@article{Asghari2025,
  title={Frailty identification using a sensor-based upper-extremity function test: a deep learning approach},
  author={Asghari, Mehran and Ehsani, Hossein and Toosizadeh, Nima},
  journal={Scientific reports},
  volume={15},
  number={1},
  pages={13891},
  year={2025},
  publisher={Nature Publishing Group UK London}
}

@article{Kim2022,
  title={Transfer learning for medical image classification: a literature review},
  author={Kim, Hee E and Cosa-Linan, Alejandro and Santhanam, Nandhini and Jannesari, Mahboubeh and Maros, Mate E and Ganslandt, Thomas},
  journal={BMC medical imaging},
  volume={22},
  number={1},
  pages={69},
  year={2022},
  publisher={Springer}
}

@article{Mustafa2021,
  title={Supervised transfer learning at scale for medical imaging},
  author={Mustafa, Basil and Loh, Aaron and Freyberg, Jan and MacWilliams, Patricia and Wilson, Megan and McKinney, Scott Mayer and Sieniek, Marcin and Winkens, Jim and Liu, Yuan and Bui, Peggy and others},
  journal={arXiv preprint arXiv:2101.05913},
  year={2021}
}

@article{Salehi2023,
  title={A study of CNN and transfer learning in medical imaging: Advantages, challenges, future scope},
  author={Salehi, Ahmad Waleed and Khan, Shakir and Gupta, Gaurav and Alabduallah, Bayan Ibrahimm and Almjally, Abrar and Alsolai, Hadeel and Siddiqui, Tamanna and Mellit, Adel},
  journal={Sustainability},
  volume={15},
  number={7},
  pages={5930},
  year={2023},
  publisher={MDPI}
}

@Article{Pal2025SLVH,
AUTHOR = {Pal, Basudha and Chellappa, Rama and Umair, Muhammad},
TITLE = {Encoding of Demographic and Anatomical Information in Chest X-Ray-Based Severe Left Ventricular Hypertrophy Classifiers},
JOURNAL = {Biomedicines},
VOLUME = {13},
YEAR = {2025},
NUMBER = {9},
ARTICLE-NUMBER = {2140},
URL = {https://www.mdpi.com/2227-9059/13/9/2140},
PubMedID = {41007703},
ISSN = {2227-9059},
DOI = {10.3390/biomedicines13092140}
}

@article{Chao2019,
title={GaitSet: Regarding Gait as a Set for Cross-View Gait Recognition}, 
volume={33}, url={https://ojs.aaai.org/index.php/AAAI/article/view/4821}, 
DOI={10.1609/aaai.v33i01.33018126}, 
number={01}, 
journal={Proceedings of the AAAI Conference on Artificial Intelligence}, 
author={Chao, Hanqing and He, Yiwei and Zhang, Junping and Feng, Jianfeng}, 
year={2019}, 
month={Jul.}, 
pages={8126-8133}
}

@article{Fan2023,
  title={Exploring Deep Models for Practical Gait Recognition},
  author={Chao Fan and Saihui Hou and Yongzhen Huang and Shiqi Yu},
  journal={ArXiv},
  year={2023},
  volume={abs/2303.03301},
  url={https://api.semanticscholar.org/CorpusID:257365830}
}

@misc{Wei2025,
      title={ViT-Linearizer: Distilling Quadratic Knowledge into Linear-Time Vision Models}, 
      author={Guoyizhe Wei and Rama Chellappa},
      year={2026},
      eprint={2504.00037},
      archivePrefix={arXiv},
      primaryClass={cs.CV},
      url={https://arxiv.org/abs/2504.00037}
}

@misc{Radford2021,
      title={Learning Transferable Visual Models From Natural Language Supervision}, 
      author={Alec Radford and Jong Wook Kim and Chris Hallacy and Aditya Ramesh and Gabriel Goh and Sandhini Agarwal and Girish Sastry and Amanda Askell and Pamela Mishkin and Jack Clark and Gretchen Krueger and Ilya Sutskever},
      year={2021},
      eprint={2103.00020},
      archivePrefix={arXiv},
      primaryClass={cs.CV},
      url={https://arxiv.org/abs/2103.00020}
}

@article{selvaraju,
   title={Grad-CAM: Visual Explanations from Deep Networks via Gradient-Based Localization},
   volume={128},
   ISSN={1573-1405},
   url={http://dx.doi.org/10.1007/s11263-019-01228-7},
   DOI={10.1007/s11263-019-01228-7},
   number={2},
   journal={International Journal of Computer Vision},
   publisher={Springer Science and Business Media LLC},
   author={Selvaraju, Ramprasaath R. and Cogswell, Michael and Das, Abhishek and Vedantam, Ramakrishna and Parikh, Devi and Batra, Dhruv},
   year={2019},
   month=oct, pages={336–359}
}

@Article{Marey2025LMIC,
AUTHOR = {Marey, Ahmed and Mehrtabar, Saba and Afify, Ahmed and Pal, Basudha and Trvalik, Arcadia and Adeleke, Sola and Umair, Muhammad},
TITLE = {From Echocardiography to CT/MRI: Lessons for AI Implementation in Cardiovascular Imaging in LMICs—A Systematic Review and Narrative Synthesis},
JOURNAL = {Bioengineering},
VOLUME = {12},
YEAR = {2025},
NUMBER = {10},
ARTICLE-NUMBER = {1038},
URL = {https://www.mdpi.com/2306-5354/12/10/1038},
PubMedID = {41155037},
ISSN = {2306-5354},
DOI = {10.3390/bioengineering12101038}
}

@article{Mollura2020,
author = {Mollura, Daniel J. and Culp, Melissa P. and Pollack, Erica and Battino, Gillian and Scheel, John R. and Mango, Victoria L. and Elahi, Ameena and Schweitzer, Alan and Dako, Farouk},
title = {Artificial Intelligence in Low- and Middle-Income Countries: Innovating Global Health Radiology},
journal = {Radiology},
volume = {297},
number = {3},
pages = {513-520},
year = {2020},
doi = {10.1148/radiol.2020201434},
note = {PMID: 33021895},
URL = {https://doi.org/10.1148/radiol.2020201434},
eprint = {https://doi.org/10.1148/radiol.2020201434}
}

@misc{Wu2019Detectron2,
author = {Yuxin Wu and Alexander Kirillov and Francisco Massa and Wan-Yen Lo and Ross Girshick},
title = {Detectron2},
howpublished = {\url{https://github.com/facebookresearch/detectron2}},
year = {2019}
}

@article{McDaniel2025,
    author = {McDaniel, Laura and Essien, Ime and Lefcourt, Samuel and Zelleke, Ephrata and Sinha, Arushi and Chellappa, Rama and Abadir, Peter M},
    title = {Aging With Artificial Intelligence: How Technology Enhances Older Adults' Health and Independence},
    journal = {The Journals of Gerontology: Series A},
    volume = {80},
    number = {7},
    pages = {glaf086},
    year = {2025},
    month = {06},
    issn = {1758-535X},
    doi = {10.1093/gerona/glaf086},
    url = {https://doi.org/10.1093/gerona/glaf086},
    eprint = {https://academic.oup.com/biomedgerontology/article-pdf/80/7/glaf086/63507694/glaf086.pdf}
}

@misc{Mendiratta2023,
  author = {Mendiratta, Priya and Schoo, Caroline and Latif, Rafay},
  title = {Clinical Frailty Scale},
  year = {2023},
  note = {StatPearls [Internet]. Treasure Island (FL): StatPearls Publishing}
}

@misc{Fan2020,
  author = {Fan, Chao and Peng, Yunjie and Cao, Chunshui and Liu, Xu and Hou, Saihui and Chi, Jiannan and Huang, Yongzhen and Li, Qing and He, Zhiqiang},
  title = {GaitPart: Temporal Part-Based Model for Gait Recognition},
  year = {2020},
  note = {Proceedings of the IEEE/CVF Conference on Computer Vision and Pattern Recognition (CVPR), pp. 14213--14221}
}

@misc{Lin2021,
  author = {Lin, Beibei and Zhang, Shunli and Yu, Xin},
  title = {Gait Recognition via Effective Global-Local Feature Representation and Local Temporal Aggregation},
  year = {2021},
  note = {Proceedings of the IEEE/CVF International Conference on Computer Vision (ICCV), pp. 14648--14656}
}

@misc{Zheng2022,
  author = {Zheng, Jinkai and Liu, Xinchen and Liu, Wu and He, Lingxiao and Yan, Chenggang and Mei, Tao},
  title = {Gait Recognition in the Wild with Dense 3D Representations and A Benchmark},
  year = {2022},
  note = {Proceedings of the IEEE/CVF Conference on Computer Vision and Pattern Recognition (CVPR), pp. 20196--20205}
}

@misc{pal2025grasp,
  author = {Pal, Basudha and Kamran, Sharif Amit and Lutnick, Brendon and Lucas, Molly and Parmar, Chaitanya and Patel Shah, Asha and Apfel, David and Fakharzadeh, Steven and Miller, Lloyd and Cula, Gabriela and others},
  title = {GRASP-PsONet: Gradient-based Removal of Spurious Patterns for PsOriasis Severity Classification},
  year = {2025},
  note = {Medical Image Computing and Computer-Assisted Intervention (MICCAI), pp. 233--243}
}

\end{document}